\documentclass[journal]{IEEEtran}
\usepackage[utf8]{inputenc}
\usepackage[english]{babel}
\usepackage{graphicx}
\usepackage[dvipsnames]{xcolor} %MCH
\usepackage{multicol} %MCH
\usepackage{amsmath}
\usepackage{algorithm}
\usepackage{algorithmic}
\usepackage{amssymb}
\usepackage{hyperref}
\hypersetup{
colorlinks=true,%linkcolor=black,
urlcolor=blue,
}

\begin{document}
\title{Identifying Population Movements with Non-Negative Matrix Factorization from Wi-Fi User Counts in Smart \& Connected Cities}

\author{Michael C. Huffman,
        Armen Davis,
        Joshua Park,
        James Curry}% <-this % stops a space
\maketitle

\begin{abstract}
Non-Negative Matrix Factorization (NMF) is a valuable matrix factorization technique which produces a ``parts-based" decomposition of data sets. Wi-Fi usercounts are a privacy-preserving indicator of population movements in smart and connected urban environments. In this paper, we apply NMF with a novel matrix embedding to Wi-Fi usercount data from the University of Colorado at Boulder Campus for the purpose of automatically identifying patterns of human movement in a Smart \& Connected infrastructure environment. 
\end{abstract}


% \fi
\IEEEpeerreviewmaketitle

\section{Introduction}
% for IEEE journal papers produced under \LaTeX\ using
% IEEEtran.cls version 1.8b and later. \cite{Savitzky-Golay}

\subsection{Urbanization}

The 20th century saw a monumental transition of human populations from rural settings into urban ones. Beginning in Europe and the United States in the 19th century, industrialization spread across the world and drove an enormous migration of people into city centers, forever altering the way we live.  In 1950, 30\% of the world population lived in an urban setting—today it is over half \cite{UN}.  Rapidly developing nations saw even more dramatic urban growth over a comparatively shorter duration. 

The 21st century will see more of the same. By 2030, the global population is expected to pass 8.5 billion, on its way to 9.8 billion by 2050. Driven in equal measures by this overall global population growth and the persistent migration into urban areas, urban centers will continue to swell. India, China and Nigeria alone are projected to add 416 million, 255 million and 189 million people respectively to their cities by 2050. By then, it is expected that Earth’s urban corridors will have added 2.5 billion people and that 7 out of 10 people on this planet will live in a city. \cite{UN} 

Today, many livability hazards characteristic of urban population centers stem from poor planning and little consideration to scalability: crime\cite{crime}, congestion\cite{traffic}, pollution\cite{pollution}, and inequity in access to services\cite{inequality} are common features of urban life and without careful consideration, will only become worse.

\subsection{Smart Cities}
Ensuring that city residents live safe, healthy, equitable and productive lives is the central focus of the Smart \& Connected Cities movement\cite{NSF}. This is a multidisciplinary push to combine big data analytics, civil engineering and city planning in order to make urban living more human-friendly. Many identified multiple areas of city life that could be improved through the union of big data and city planning: decentralized energy production and distribution\cite{imbault}, integrated business/economic models that anticipate a demand for goods and services\cite{industry}, optimized transportation minimizing pollution and congestion\cite{transportation}, and real-time warnings with enhanced emergency-response\cite{emergency}. At its core, a central premise of Smart Cities is an intelligent response to accurate and real-time mobility data.

The promises of Smart Cities to improve the human condition can only be realized through widespread data collection and sophisticated understanding of the movements of human population through their environments. This is a critical and active area of research which capitalizes on the growth of personal connected devices that are so common today. A 2016 study of population movement in the industrial cities of northern Italy demonstrated the insights possible from telecommunications data \cite{Khodabandelou}. Another 2015 study of telecommunications data demonstrates the power to reveal patterns of human mobility \cite{Meyer}. However, such approaches have significant drawbacks. Telecommunication data may only be available to carriers and not the municipalities. Above all, such surveillance techniques are considered intrusive and come with major privacy concerns\cite{privacy}. 
%Energy efficiency of buildings can be improved by as much as 30%\$
\subsection{Wi-Fi}

Wi-Fi describes a family of wireless communication protocols which operate over unlicensed radio frequencies and allow devices to be networked in small scale environments (homes, offices, and shops). Introduced in 1999 and governed by the family of standards IEEE 802.11\cite{ieee802}, today Wi-Fi access is a common feature of urban environments. With the growing popularity of public hotspots, Wi-Fi expected to see exponential growth in the coming years. According to a 2018 industry white paper by telecommunications giant Cisco Systems, Inc., public Wi-Fi hot-spots are expected to increase to 628 million globally by 2023\cite{cisco}--a fourfold increase over 5 years. 

Wi-Fi has been previously identified as a valuable analogue for population data. Researchers have shown that Wi-Fi can produce high fidelity population counts of buildings\cite{wang}, even more accurately than the HVAC industry standard of CO$_{2}$ sensors\cite{Ouf}.  Building on this success, Wi-Fi has been identified as a method to monitor and forecast crowds in large-scale public events\cite{Singh}. Other teams have validated Wi-Fi as a method of assessing city behavior with real-world data\cite{Bellini} and there's even been work done to develop a real-time city census from Wi-Fi data\cite{Kontokosta}.

In addition to its accurate representation of human populations, by only collecting active user counts at Wi-Fi access points the technology can be rendered anonymous and deployed as a privacy-preserving mode of monitoring population movement. Because of these attractive features--accuracy, ubiquity, growth, and anonymity--it is the belief of the authors that Wi-Fi is a safe tool to monitor population movements and will play a critical role in realizing the lofty goals of Smart Cities.

\subsection{Matrix Decomposition of Data}

 Unsupervised, data-driven approaches to automatically extract a low number of latent components from complex data are of high demand in signal separation problems such as  single-channel audio source separation or hyperspectral unmixing. Indeed, there is a similar problem present in the Wi-Fi user counts of Smart \& Connected infrastructure. The Smart \& Connected Cities movement hinges on our ability to understand and predict human movement through an urban environment and respond intelligently. With that in mind, a natural question is how can bulk movements of people and their devices into and out of buildings be identified from single-channel signal of population data?

 %\textbf{Add stuff here about history of NMF as well as the reason behind such a decomp. See pg. 12 from NMF textbook}
 
 The canonical way to do such a reduction is the singular value decomposition (SVD). With similar intentions, we stray from this path in favor of the Non-negative Matrix Factorization (NMF). Unlike SVD, NMF yields strictly non-negative components of the data. While both factorizations seek similar results in the form of, for example, dimensionality reduction \cite{NMFTextbook}, NMF is thought to return more interpretable results since many physical quantities are constrained to be positive or non-negative (e.g. atmospheric pressure, user counts). For these data, standard SVD will, in general, return nonphysical components of the original data matrix which leads to interpretability issues. Due to the non-negativity constraint imposed by NMF, the components can be thought of as additive parts used to reconstruct the data \cite{Lee}. Since Wi-Fi user counts are strictly non-negative, NMF is certainly a compelling alternative to SVD.
 
 NMF has a rich history dating back to the early 1960s in the fields of analytical chemistry and remote sensing \cite{NMFTextbook}. The work done in the 60s was not directly referred to as NMF, although it was equivalent. In fact, NMF did not appear in its present form until 1994 when a paper was published on \textit{positive matrix factorization} \cite{PosMatFact}. The true explosion of the topic can be directly traced to the \textit{Nature} article by Lee and Seung in 1999 when data compression and feature extraction were shown to be intrinsic properties of NMF \cite{Lee}.

\section{Methods and Materials}

\subsection{University of Colorado Boulder Wi-Fi Data}

Data for this study was provided by the Office of Information Technologies (OIT). We generate total active user counts on a per-building basis by consolidating multiple Wi-Fi access points (APs) that exist in each building. The access points of the school are designed to be used in a mesh-like mode where many access points provide coverage for a large area, while devices ebb and flow between them. As a consequence, a device is only ever counted by one AP and duplicate counts for a building are avoided.

Unfortunately the University of Colorado Boulder (UCB) system's sampling frequency is almost never constant and we have no control over when the system opts to report device counts. User count sampling intervals are therefore irregular, and vary across time and space. Sampling intervals generally range from 3-12 minutes, and irregular outages in data collection can be hours or days. 

By taking the connected counts throughout the day we obtain time series data which will be analyzed by NMF. The data sets analyzed here have been stripped of all identifying information leaving only active user counts. As mentioned earlier, this is one of the most attractive features of this dataset: complete anonymity with no privacy rights breached. In the Results section of this paper, we introduce and study the Norlin Library.

\subsection{Data Pre-processing and Matrix Structure}

%I plan on combining JP's sections below (D and E) into one section describing all pre-processing stuff. 

\textit{Interpolation}\\
As with many time series techniques, it is helpful to have a standard sampling frequency of the data for interpretation reasons. Non-constant sampling frequencies would throw a wrench into the construction and interpretation of the data matrix $X$. We settled upon ten-minute interpolation of the data although, admittedly, this was a bit arbitrary. Due to the objective of understanding general, aggregate behavior, we decided to perform a \textit{linear} interpolation as opposed to quadratic, cubic, etc.. We left all fractional interpolated counts unchanged for the analysis since the NMF components will not, in general, be whole numbers. 

% \textit{Bad Data}\\ 
% Identifying and addressing corrupted or missing data is a crucial step in all real-world data science. 

\textit{Matrix Embedding}\\
Although factorization techniques act on matrices, most time series data is typically one-dimensional. Therefore, a critical choice for this type of analysis is how to impart 2-dimensional structure to the time series. Finding ways to embed this data into a 2-dimensional matrix is crucial in terms of the final product. Namely, how $W$ and $H$ are interpreted and how well the product $WH$ approximates $X$ vary with the design of $X$. 

In other time series approaches such as Singular Spectrum Analysis \cite{SSA}, data is formed into a \textit{Hankel matrix}, where each column consists of lagged vectors. In our experience, NMF via a Hankel matrix interpretation has applications in forecasting for streaming data (though there is an argument to be made about information loss during Hankelization procedures). Instead, in a further attempt to extract "real" meaning, e.g. features from our factorization, we seek to mirror the daily cycle of building use by choosing to represent each 24 hour period of data collection as a column of the design matrix, $X$. 
This embedding is not without precedent: for large parts of the procedure we follow the approach in this paper by Lee and Staneva \cite{Echo-Sounder} by treating each day of data as a sample. This is justified by the clear daily oscillation in Wi-Fi patterns coming from general group movements around campus. This leads to a knowledge based factorization.

Using the ten-minute interpolation, we obtain 144 data points per day. Each day is treated as a sample, and they appear chronologically as a columns in $X$. Thus, the factors $W$ and $H$ have simple interpretations. Namely, the columns of $W$ represent daily patterns and the rows of $H$ represent the weights of the daily patterns for each day in the original data. This is where the importance of constant sampling frequencies comes into play. Having the columns of our data matrix referring the same times of the day is crucial for a consistent comparison. Further, the columns would likely be of different dimension without interpolation.

This embedding helps constrain the factorization and optimizes NMF's ability to produce results with physical interpretability. Given any day in the training data, we can reconstruct its pattern via a linear combination of the columns of $W$ where the coefficients (weights) are represented by a column of $H$ corresponding to the desired day. It is also worth noting that we can examine the prominence of each column of $W$ on any given day by examining the weights in the $H$ matrix. 
\subsection{Choice of Inner Dimension}

Determining optimal choices for the dimension of the truncated SVD,  $r$,  has been an outstanding issue for full-rank matrices. The same can be said for the choice of inner dimension, $k$, in NMF since both parameters seek similar results in terms of dimension reduction. For certain matrices, optimal hard-thresholding techniques have been developed for SVD, but more-informal strategies such as examining bends/elbows in a plot of decreasing singular values are still used today \cite{DataDriven}. We use a similar strategy for the determination of $k$ via the graphical examination of mean-squared-error (MSE) vs. $k$. 

We look for the presence of sharp bends/elbows in this plot of MSE vs. $k$ using a sufficiently wide range of $k$ values. The entire NMF procedure is re-computed for each successive value of $k$ and MSE is computed in the obvious, element-wise manner between the resulting product $WH$ and $X$. 

When in doubt, we advocate for potentially underestimating $k$ rather than risking an overestimate. No matter what $k$ is specified to be, the NMF algorithms \textit{will} make factors $W$ and $H$ that fit the specified dimensions. If $k$ is overestimated, we risk extracting components that are numerical artifacts of the algorithm rather than true components of the data. If $k$ is underestimated, then it is more likely that the components extracted are truly parts of the data, or contain several hidden parts of the data. We would rather see a few, true components of the data rather than many components that might be a mix of truth and artifact.

\subsection{NMF Details}

The problem setup for NMF is as follows. Given a data matrix, $X \in \mathbb{R}^{n\times m}$, we seek factors $W \in \mathbb{R}^{n \times k}$ and $H \in \mathbb{R}^{k \times m}$ (where the so called inner dimension $k \ll min\{n, m\}$) having the properties

\begin{equation}
    X \approx WH \;\;\;\;\;\;s.t. \; W, H\succeq0
    \label{eq:NMF}
\end{equation}

where $W, H\succeq0$ denotes element-wise non-negativity. Similar to its relative, SVD, dimension reduction is usually an assumed goal. The alternative $X \approx WH^T$ is commonly used to describe the same factorization, but we follow the notation of \cite{NMFTextbook}. 

This factorization, although easy to describe, is quite difficult to solve in general. Not only is it fundamentally ill-posed and non-convex, but it is NP-hard in all but a very few specific cases \cite{NMFTextbook}. With real data, the NP-hardness is essentially guaranteed as exact low-rank structure rarely exists.  

We note that \textit{many} variants exist in terms of how $W$ and/or $H$ are constructed. Twenty variants are listed and discussed in-depth in \cite{NMFTextbook}. For example, \textit{deep-NMF} seeks $X \approx WH_1H_2H_3...H_t$ where $W, H_i\succeq0$ $\forall i$ and \textit{projective-NMF} adds the additional constraint $W = XH^T$, $H\succeq0$. The style of NMF chosen is generally application-dependent. In this paper, we stick to the traditional form of NMF given by (\ref{eq:NMF}).

In general, no perfect factors are guaranteed to exist with NMF (in contrast to SVD), which is a consequence of the non-negativity constraint (hence the``$\approx$"). Therefore, NMF is set up as an optimization problem. For NMF, there are at least four critical choices that are associated with any optimization problem: Objective function, regularization terms, descent algorithm and algorithm initialization. To give a more comprehensive look at NMF, we will cover the objective function and regularization terms first and follow with the algorithm and  initialization procedure.

\subsubsection{Objective Function}

Traditionally, NMF seeks 

\begin{equation}
    \operatorname*{argmin}_{W\succeq0, H\succeq0} \left||X-WH \right||_F^2
    \label{eq:tradiationalOptimization}
\end{equation}

where $||\cdot||_F$ denotes the Frobenius norm. There is the choice of many different objective functions that differ from the Frobenius norm, however. The general objective function can be seen as the $\beta$-divergence, with some choice of $\beta$ \cite{NMFTextbook}. Although, we refer to reader to $l_1$-norm, $l_{\infty}$-norm and weighted versions of (\ref{eq:tradiationalOptimization}) seen in \cite{NMFTextbook}. We will not cover these alternatives. The $\beta$-divergence between two non-negative scalars $z$, $y$ is given by

\begin{equation}
    d_{\beta}(z, y) = 
    \begin{cases} 
      \frac{z}{y} - log(\frac{z}{y}) - 1 & \beta = 0\\
      zlog(\frac{z}{y}) - z + y & \beta = 1\\
      \frac{1}{\beta(\beta-1)}\left(z^{\beta} + (\beta-1)y^{\beta} - \beta zy^{\beta-1}\right) & \beta \neq 0, 1
   \end{cases}
   \label{eq:betaDiv}
\end{equation}

The three most common objective functions correspond to $\beta = 0, 1, 2$. In other words, the Itakura-Saito divergence (IS-divergence), the Kullback-Leibler divergence (KL-divergence) and the Frobenius norm, respectively. For our purposes, we chose the Frobenius norm ($\beta = 2$) due to it successes in Blind Source Separation (BSS) \cite{sawada}. 

\subsubsection{Regularizations}

It is also possible to regularize the problem given in (\ref{eq:tradiationalOptimization}) no matter the objective function. Regularization induces certain solution types (e.g. sparsity) and is often used to make the solution to the optimization problem unique. For our purposes, we use the regularization option that is included in the NMF function in Python and is part of the Scikit-Learn package \cite{sklearn}.
Thus, with $\beta = 2$, the optimization problem becomes

\begin{equation}
    \begin{split}
    \operatorname*{argmin}_{W\succeq0, H\succeq0} \frac{1}{2}\left||X-WH \right||_F^2 + \\
    \rho\alpha\left( \left||\text{vec}(W)\right||_1 + \left||\text{vec}(H)\right||_1\right)  + \\ 
    \frac{1}{2}\alpha(1-\rho)\left(\left||W\right||_F^2 + \left||H\right||_F^2\right) 
    \end{split}
    \label{eq:fullOptimizationProblem}
\end{equation}

where $\alpha$ controls the intensity of the regularization and $\rho$ controls the relative importance of element-wise $l_1$-norms and the Frobenius norms. As a potentially useful comparison, we note that (\ref{eq:fullOptimizationProblem}) has many similar properties to the elastic-net regression problem.

\subsubsection{Algorithm} There are two approaches to solve the problem given in (\ref{eq:tradiationalOptimization}) and (\ref{eq:fullOptimizationProblem}). The first is the classical \textit{gradient descent} method. With linear problems gradient descent is usually straightforward to implement since partial derivatives exist and can easily be found. For example (\cite{HALS}) showed that the gradient descent algorithm is equivalent to the more widely-used \textit{multiplicative update} strategy, where each matrix is updated geometrically as opposed to the arithmetic update of gradient descent. 

A second widely-used strategy stems from recognizing that the problem given in (\ref{eq:tradiationalOptimization}) is actually convex once either $W$ or $H$ are known. Given one above, classical least squares can be used to solve for the other unknown. Initializing $W$ or $H$ then implies a procedure where classical least squares optimizes the other matrix during each iteration. This approach is called \textit{alternating least squares} and it is the basis for the \textit{coordinate descent strategy}.

\subsubsection{Initialization Strategy}
%Talk about randomization, svd, nndsvd and its variants, our choice of nndsvdar 

Regardless of algorithm choice (multiplicative update or coordinate descent), we must specify an initial $W$ and/or $H$ to start the procedure. Built-in NMF solvers in most scientific computing languages (e.g. Scikit-Learn's NMF function for Python, MATLAB's standard \textit{nnmf()} function) initialize with random matrices at default. So, each entry in $W$ and/or $H$ is drawn from a $Uniform(0, 1)$ distribution. Due to the non-convex nature of the problem and the large number of unknowns (every entry in $W$ and $H$), random initialization often results in slightly poorer outcomes and reproducibility issues. We advocate for limiting the stochasticity of the algorithm, in favor of more reproducible results via a SVD initialization strategy. 

Namely, we use the $nndsvdar$ call within the Scikit-Learn NMF function. This initialization comes from the non-negative double SVD (NNDSVD) strategy proposed in Boutsidis et al.\cite{NNDSVD}. We remind the reader that, for \textit{any} matrix $A \in \mathbb{R}^{m \times n}$, there exists a decomposition

\begin{equation}
    A = U\Sigma V^T
\end{equation}

such that $U \in \mathbb{R}^{m\times m}$, $\Sigma \in \mathbb{R}^{m\times n}$ and $V \in \mathbb{R}^{n\times n}$. The columns of $U$ are the eigenvectors of $AA^T$, the columns of V are the eigenvectors of $A^TA$, and $\Sigma$ is rectangular diagonal matrix filled with the singular values. For many dimension reduction problems, \textit{truncated} SVD is used to form a rank-$r$ approximation to the original matrix, $A$, via the Eckart-Young Theorem \cite{DataDriven}.

Standard non-negative double SVD (NNDSVD) examines a rank-$r$ approximation of the original data matrix. The choice of the inner dimension, $k$, for NMF is the same  as $r$ for the truncated SVD. It then examines the positive and negative components of each column of $U$ and row of $V^T$ and determines which has a larger contribution (by comparing their norms) to each of the singular vectors. The branch (positive or negative) with the larger contribution is then chosen to replace the column of $U$ or row of $V^T$ in its absolute value. The singular values are square-rooted and equally distributed between the the $U$ and $V$ matrices. These scaled, non-negative matrices $U$ and $V^T$ are then treated as the initialization to $W$ and $H$, respectively. See  Boutsidis and Gallopoulos \cite{NNDSVD} for the full algorithm and discussion. The difference between this strategy and the $nndsvdar$ call is that any values in the NNDSVD initialization equal to zero are replaced with small, random values. The reason behind this is to allow zero entries to be updated in the descent algorithm. In their basic forms, both coordinate descent and multiplicative updates cannot alter zero entries.

%The general problem setup (variant of NMF, initialization, objective function, regularizations and algorithm)  are decided based upon the data used, the construction of $X$ and the desired outcome of the components. 

\section{Results}
\begin{figure*}[h!]
    \centering
    \includegraphics[width = \textwidth]{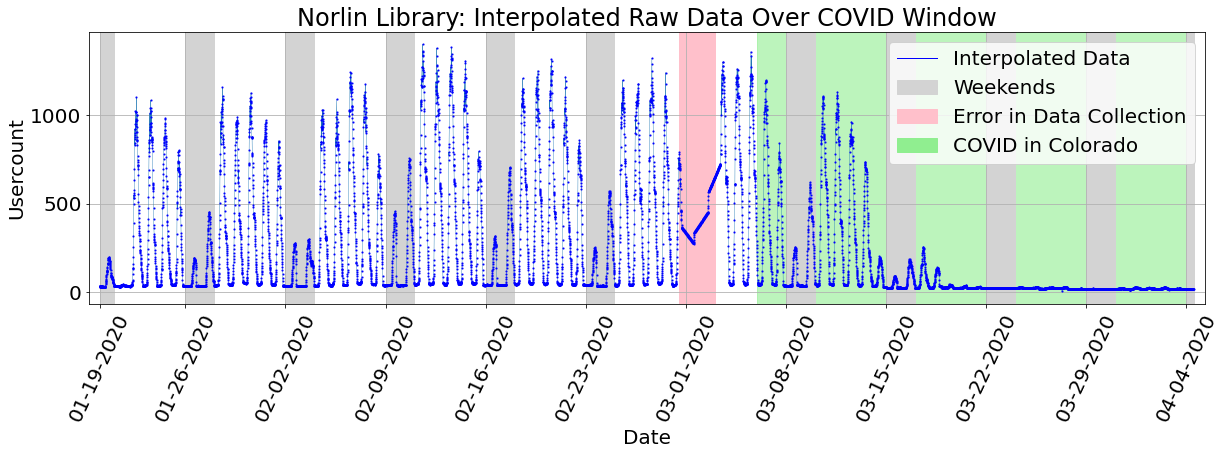}
    \caption{\textbf{Norlin Library Timeseries: raw data from the Spring of 2020}- A large, unabridged data set was studied with the NMF technique described above. Prominent features of this timeseries include strong daily cycles, weekly variation, a large subset of corrupted data, and the beginning of the ``COVID effect" in Colorado.}
    \label{NorlinData}
    \centering
    \includegraphics[width = \textwidth]{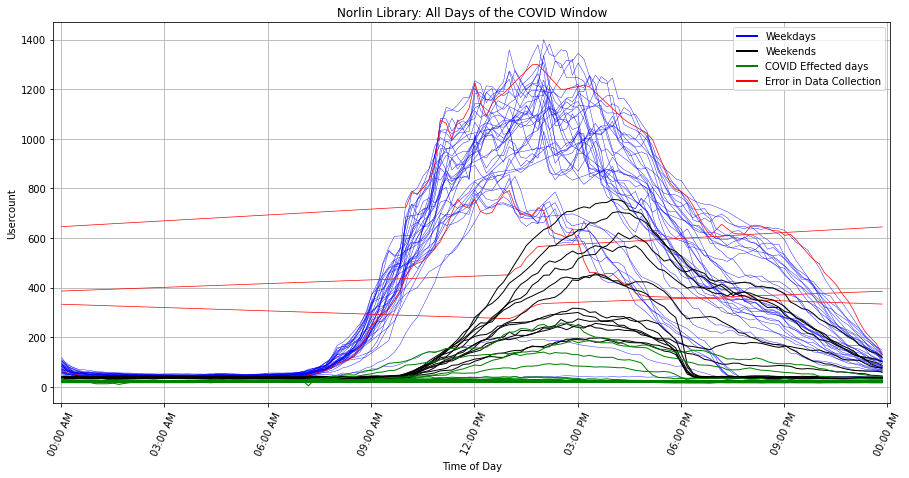}
    \caption{\textbf{$X$: All Days of the Study} - The same timeseries data as above shown in a daily view. Alternatively, this could be considered columns of the data matrix, X. Note the same types of variation as above: Weekends, Weekdays, Error in Data and COVID.}
    \label{AllDays}
\end{figure*}
\subsection{COVID Effect: Norlin Library}
To illustrate this method’s ability to identify structure in a highly varied dataset, it was applied to UCB’s main library: Norlin Library. This study was carried out over an 11-week period from the Spring of 2020 spanning the (reported) beginning of the COVID-19 epidemic in Colorado\cite{coloradocovid}. With regard to the setup discussed in the previous section, our data matrix $X \in \mathbb{R}^{n\times m}$, where $n = 144$ (the number of sampling intervals in a day) and $ m = 77$ (77 unique days in this study). 
\begin{figure*}[h!]
    \centering
    \includegraphics[width = \textwidth]{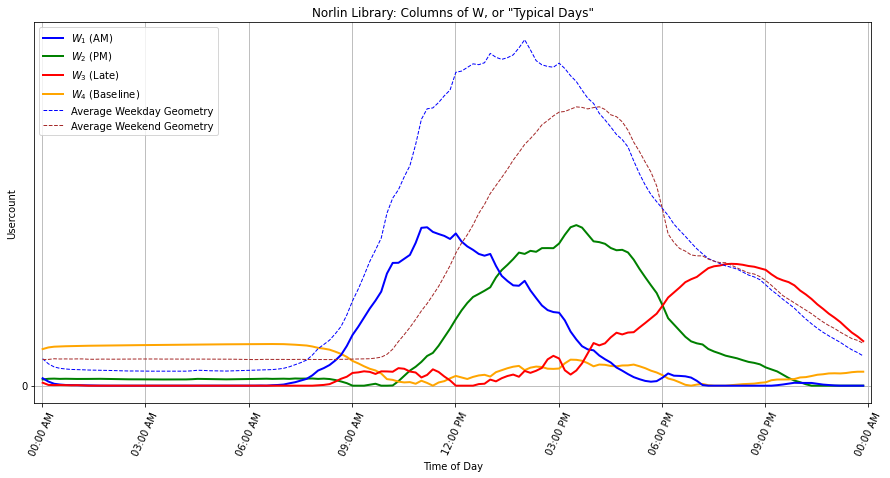}
    \caption{Norlin Library: Columns of W or ``Typical Days". After NMF with $k = 4$ (see justification in Results), these are the extracted basis vectors by which the algorithm represents every day in the data set. Also in this figure are the ``Average geometry of Weekdays" and ``Average Geometry of Weekends". Keep in mind all days are represent by some linear combination of these 4 signals. We've named the columns, $W_i$, based on when they represent users in the library. NOTE: There is no scale on the y-axis scale in this figure, as the magnitudes of these curves is somewhat arbitrary.}
    \label{NorlinW}
    
    \centering
    \includegraphics[width = \textwidth]{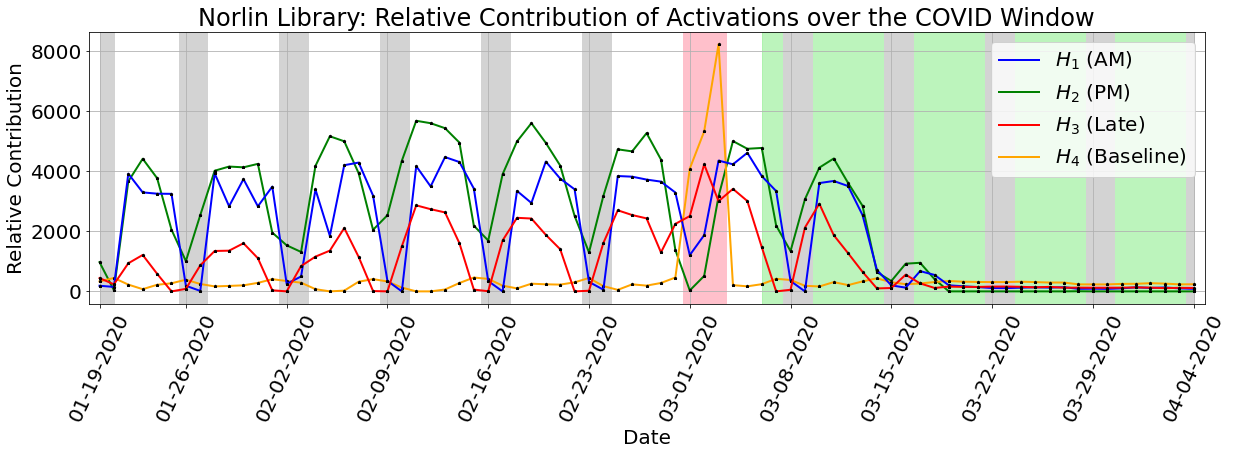}
    \caption{\textbf{$H$: Activation Contribution to each day's profile} - Here are the activations for each day multiplied by the L1 Norm of their corresponding Column of W. In this way, we can compare the relative signal strengths of columns of $W$. Do these results make physical sense?}
    \label{NorlinH}
\end{figure*}
Norlin Library is a major building located north and central on CU’s Boulder main campus. It houses the Norlin “Stacks”—-a major collection of humanities, social science and life science documents, and rare volumes-— and includes diffuse study areas and a coffee shop. 
\begin{figure}[h!]
    \raggedright
    \includegraphics[width=\columnwidth]{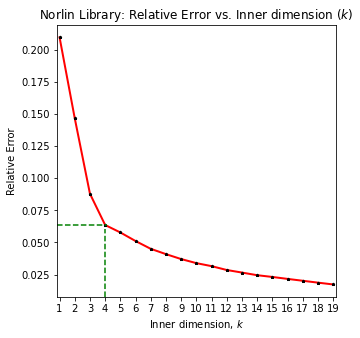}
    \caption{What inner dimension to choose? Relative error vs. Inner dimension for the NMF factorization of the Norlin Library Study Data strongly supports the choice of $k = 4$ in this study.}
    \label{NorlinK}
\end{figure}
Because very few if any classes are taught at Norlin, the building behaves more as a typical commercial building: there are fewer coordinated movements of large crowds into and out of the building. Since there are no residents, a stronger daily structure is observed in the user count data than at the residential halls. At night, the building is empty aside from the baseline count of latent Wi-Fi users (networked hardware such as vending machines, web cameras, and permanent computers). In this way, the flow of humans (and their devices) might better approximate the surge typical of commercial buildings in Smart \& Connected Cities. 

This study began on 01/19/2020 and ran for 11 weeks through 04/04/2020. In this study, the data was not  segregated into “like” days nor were outliers or corrupted data removed. Rather, the entire unabridged interval of this time series was passed wholesale to the algorithm. 

Before discussing the results of the factorization, let's take a closer a look at the original time series data. This interval demonstrates some intriguing features—some regular, some irregular—that make it an insightful case study of the techniques described in this paper. Some major ones to note, which are also highlighted in Figure \ref{NorlinData} and Figure \ref{AllDays}: 
\begin{itemize}
    \item Weekdays vs. Weekends: User count populations demonstrate much different structure from weekdays to weekends.
    \item Corrupted Data: for a roughly 3 day period from 3/01 to 3/03, the data is obviously corrupted and does not reflect the true behavior of devices in the library. Typically, many techniques would require operator intervention to throw out this data. Here it is included to demonstrate the robustness of this technique. 
    \item COVID-19: Perhaps most conspicuous is the obvious “paradigm” shift from the pre-COVID era to the COVID era in Colorado. On March 5th, the Governor of the state announced the first confirmed case of COVID in Colorado. Shortly thereafter the user count data reflects the steady de-peopling of campus. 
\end{itemize}
Per the procedure prescribed in the previous section, this time series data was vectorized by day, loaded into a data matrix and factorized. As before, a variant of Hierarchical Alternating Least Squares (HALS) algorithm was used to solve the factorization, and initialized by a non-zero SVD method. For this application, a relatively low inner dimension of $k=4$ was chosen and this rank-selection is supported by the results of an MSE analysis (Figure \ref{NorlinK}).

The resulting factorization successfully found a representation of the data by parts that seems to reflect plausible structures in the movement of device populations. The 4 columns of $W$ representing the constituent signals of this system are shown in Figure \ref{NorlinW}. They have been assigned descriptive names based on when they represent devices in the Library: $W_{1}$ (AM), $W_{2}$ (PM), $W_{3}$ (Late) and $W_{4}$ (Baseline). Keep in mind that the NMF has “learned” a representation of every day in our data set as a linear combination of these 4 vectors. 
Examining the weight matrix, $H$ gives some insight into the temporal structure of user counts in the Library. In Figure \ref{NorlinH}, we show each of the 4 weights for every day in the times series. These weights are scaled by the Manhattan distance of their corresponding columns of W, or $W_{j}$:
\begin{equation}\label{weighting_h}
    h_{i, j, weighted} = h_{i, j} \times \left \| W_{j} \right \|_1 \times \text{10 min}
\end{equation}
This scaling is performed to normalize the weights for comparison and impart a physical intuition to them. $\left \| W_{j} \right \|_{L_{1}}$ is an effective measure of cumulative device time each day or  when multiplied by  10 minutes can be considered more directly a “device minutes per day” value (due to the imposed 10 minute sampling interval) represented by each column of $W$ ($W_{j}$). Numerical integration may provide a more refined value, but either is sufficient for understanding the relative contributions of each column of $W$ for each date. This new value $h_{i, j, weighted}$ therefore allows us to directly compare the “signal strengths” of each $W_{j}$ relative to one another, and more critically, how those relationships change temporally. Several patterns can be identified:
\begin{itemize}
    \item $W_{1} (AM)$ is strongly active on weekdays and essentially “off” on weekends. This signal could correspond to the ``traditional crowd" that only comes to campus during weekdays. 
    \item $W_{3} (Late)$ shows a similar pattern, though curiously is offset a day (“off” on Friday and Saturday). This signal could correspond to the ``homework crowd", i.e. those individuals who take a break from class and homework at the end of the week but return Sunday night to do work before the coming week. 
    \item The entire period of error in data collection has a drastically different profile and stands out clearly from anywhere else on the plot. $W_{4}$ sees a 40-fold increase in expression compared to anywhere else in the sampling period, while $W_{2}$ sees a unique reduction. It is an interesting finding that NMF can detect these types of errors unsupervised. 
     \item As COVID begins, the signals for $W_{1}$ (AM), $W_{2}$  (PM), $W_{3}$  (Late) decay as the campus empties, but the magnitude of $W_{4}$ actually stays constant around 450 device hours per day. This corresponds to approximately 20 devices on the network day, which indeed appears to be the background device count.
\end{itemize}

\section{Conclusion}

NMF was chosen as a decomposition for this data due to its high interpretability. As discussed earlier though, there are no guarantees made on the uniqueness or indeed suitability of the results of NMF factorization. Rather, results must be analyzed and interpreted to confirm physical meaning. Separating the data from the time series into daily vectors is a natural way to frame our factorization and imparts a physical intuition to the problem.

In the Norlin Library study, a complex, real-world data set was passed through the algorithm described in this paper. Despite being highly varied (including corrupted data), the resulting factorization found structure in the population time series that reflected the real world use of the library.

Important next steps would be to study how increasing the choice of inner dimension of the factorization may identify finer features in the population data or using a separate data source (such as transportation data) to identify specific instances of coordinated arrivals or departures into the Wi-Fi environment.

We are excited by the techniques described and believe they may be deployed to similar use cases in order to quickly, automatically and anonymously find patterns in population movements within Smart \& Connected environments.

% or


\section*{Acknowledgments}

Thanks to OIT and the Campus Chief Security Officer for making Wi-Fi data available. Many of the ideas and framing were developed through interactions with colleagues in the CU Boulder Technology, Cybersecurity and Policy (TCP) program, e.g Dan Massey, Kevin Gifford, Keith Gremban, and Lloyd Thrall. Seminar presentations and speakers had a formative impact on this work. Thanks to Peter Borrell and Jake McGrath for many valuable formative discussions. Finally, special thanks to Sandia National Laboratory for the support of this research project.

\ifCLASSOPTIONcaptionsoff
  \newpage
\fi

\end{document}